\documentclass{article}
\usepackage{icmc2018template}
\usepackage{times}
\usepackage{ifpdf}
\usepackage{soul}
\usepackage[english]{babel}
\usepackage{array,multirow}
\usepackage{caption}
\def\papertitle{Mugeetion: Musical Interface Using Facial Gesture and Emotion}
\def\firstauthor{Eunjeong Stella Koh}
\def\secondauthor{Shahrokh Yadegari}
\newif\ifpdf
\ifx\pdfoutput\relax
\else
   \ifcase\pdfoutput
      \pdffalse
   \else
      \pdftrue
  \fi
\fi

\ifpdf % compiling with pdflatex
  \usepackage[pdftex,
    pdftitle={\papertitle},
	pdfauthor={\firstauthor, \secondauthor},
    bookmarksnumbered, % use section numbers with bookmarks
    pdfstartview=XYZ % start with zoom=100% instead of full screen; 
   ]{hyperref}
  \usepackage[pdftex]{graphicx}
  \graphicspath{{./figures/}}
  \DeclareGraphicsExtensions{.pdf,.jpeg,.png}

  \usepackage[figure,table]{hypcap}

\else % compiling with latex
  \usepackage[dvips,
    bookmarksnumbered, % use section numbers with bookmarks
    pdfstartview=XYZ % start with zoom=100% instead of full screen
  ]{hyperref}  % hyperrefs are active in the pdf file after conversion

  \usepackage[dvips]{epsfig,graphicx}
  \graphicspath{{./figures/}}
  \DeclareGraphicsExtensions{.eps}

  \usepackage[figure,table]{hypcap}
\fi

\hypersetup{
    colorlinks,%
    citecolor=black,%
    filecolor=black,%
    linkcolor=black,%
    urlcolor=black
}

\title{\papertitle}
\author{\firstauthor}
\twoauthors
  {1.5in}
  {\firstauthor} {Music Department \\ UC San Diego \\ %
     {\tt \href{mailto:author@unt.edu}{eko@ucsd.edu}}}
  {\secondauthor} {Music Department \\ UC San Diego \\ %
     {\tt \href{mailto:author@unt.edu}{sdy@ucsd.edu}}}

\begin{document}
\capstartfalse
\maketitle
\capstarttrue
\begin{abstract}
People feel emotions when listening to music. However, emotions are not tangible objects that can be exploited in the music composition process as they are difficult to capture and quantify in algorithms. We present a novel musical interface, Mugeetion, designed to capture occurring instances of emotional states from users' facial gestures and relay that data to associated musical features. Mugeetion can translate qualitative data of emotional states into quantitative data, which can be utilized in the sound generation process. We also presented and tested this work in the exhibition of sound installation, Hearing Seascape, using the audiences' facial expressions. Audiences heard changes in the background sound based on their emotional state. The process contributes multiple research areas, such as gesture tracking systems, emotion-sound modeling, and the connection between sound and facial gesture.

\end{abstract}

\section{Introduction}\label{sec:introduction}

Electronic music researchers use various components as inputs for their music generation process \cite{poupyrev2001new,lyons2017machine,lyons2001facing,ccamci2012cognitive}. Music and emotion are strongly linked, and listeners can feel different emotions directly or indirectly through music. Engaging emotion as a component of a musical interface has great potential for composing creative music and expressing messages in an effective way \cite{ventura2009emotion}. However, there are several difficulties in using emotion for sonification \cite{leman2008embodied,winters2013sonification}. First, emotion is qualitative and thus hard to utilize for sound generation applications, which rely on quantitative inputs. Second, emotion is represented on a continuous spectrum. Measurement of affect requires a complex and multi-faceted approach. In this paper, we use a facial gesture tracking system to define emotional states based on facial gesture information.

Facial gestures express various information related to emotion, cognition, and inspiration \cite{du2014compound,kanade2000comprehensive}. Further, facial gestures are more straightforward indicators of emotion than other bodily gestures. There are several studies related to the connection between facial gestures and sound itself \cite{d2013mageface,faceproject}. In this paper, we propose an interactive audio interface that sonifies emotion. The idea is to use facial gesture data to detect emotion and categorize these into several emotional states for sonification. We implement two approaches for this prototype: (1) music style transition based on user's emotion and (2) auditory interface based on the connection between facial components and musical metadata. We also installed our system in a digital exhibition for facial interaction with an audience at the \textit{Hearing Seascape} installation. During the exhibition, Mugeetion detected audience's facial expression in real-time and audience were able to hear the sounds simultaneously which was mapped with their specific facial gestures.

\section{Backgrounds and Related Work}

There has been a rich history of creating novel sound interfaces using gesture-based motion-tracking for live performance and improvisation \cite{kussner2014musicians,jensenius2012motion,churnside2011musical}. A motion tracking system can allow a musician to generate their own creative music in real-time \cite{wang1998hybrid,liu2015video}. Previous studies demonstrated interesting new audio interfaces for sonification through body gesture. A number of systems have looked at capturing gestures and utilizing gesture data for the sonification process either stepwise or in real-time \cite{churnside2011musical,migicovsky2014moveosc}. Regarding previous studies, there are two approaches, which have used sound as an input for tracking facial gestures or facial components as input for sound generation. Some studies utilized auditory input for focusing on the visualization of facial gesture \cite{kramer2000auditory,sedes2004visualization,hawkinsproject}. For example,  Kapuscinski \cite{Kapuscinskiproject} conducted listening tests of Chopin pieces and recorded facial expressions from the participants. Other experiments have focused on sound generation using facial parameters as an input \cite{youtubefaceosc}. These studies use FaceOSC software to apply facial gesture data to the sound generation process. McDonald \cite{faceproject} created FaceOSC software to track facial gestures directly to Max as input.
% For example, d’Allessandra, Atstrinaki....... 
There are several interesting experiments linking facial gestures and sound on Youtube \cite{youtubefaceosc}. %\footnote{\url{https://www.youtube.com/results?search_query=faceosc}}. 
However, these experiments are more targeted toward application, rather than music cognition research.  Few computer music researchers delve into the relationship between emotion and sound itself. 

Music psychologists have studied the relationship between emotion and sound, and tried to model its connection. However, music cognition research has not contributed to music sonification research. In this paper, we propose a musical interface with the facial gesture tracking system and Facial Action Coding System (FACS) \cite{du2014compound} in order to capture emotional states. FACS can allow a concrete data representation of the facial gesture and its corresponding emotional state. Thus, facial gestures can be reference points for observing emotion and translating emotion into sound. In this context, we use the tracked facial data to the sonification process. 
We will discuss its musical implementation in the following sections. 

% \subsection{Gesture Tracking System and Audio Interface}

% The background of this work comes from the field of video image processing and motion tracking system \cite{foxlin2001motion}. The development of these studies has made turning motion into sound possible \cite{kussner2014musicians}. A number of systems have looked at capturing gestures using motion tracking systems, which are surveyed in \cite{churnside2011musical,migicovsky2014moveosc}, and utilized gesture data for their sonification process. 

% % \subsection{Music and Emotion}
% In music cognition area, there have been a lot of studies for designing the model between emotion and sound. They strongly argued that emotion and sound have strong relationship based on several reasons. %%

% % \subsection{Facial Gesture, Emotion and Sound}

% This experiment has been implemented based on Max/MSP workspace \cite{puckette1998real}. FaceOSC allows user to utilize these data for processing (Figure 2), such as mouth width/height and eye gesture.

\section{Methods}
\subsection{Understanding Facial Gesture}

\begin{figure}[h]
\centering
\includegraphics[width=8cm, height=4cm]{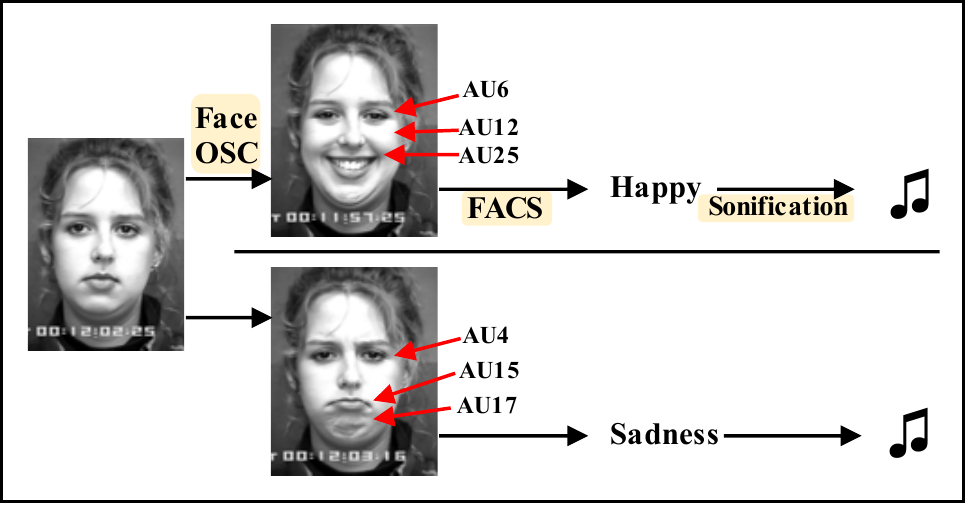}
  \caption{System structure: connection between facial gesture, compound facial expressions of emotion, and sound}
  \label{fig:system_structure}
\end{figure}

Figure \ref{fig:system_structure} gives an overview of integrating the proposed system to connect facial expression to sound generation\footnote{In Figure 1 and 2, printed images are copyrighted by \raisebox{.5pt}{\textcircled{\raisebox{-.9pt} {c}}}Jeffrey Cohn, which come from Cohn-Kanade (Ck \& CK+) database. \url{http://www.consortium.ri.cmu.edu/ckagree/}}. We generated the musical style based on facial expressions. In Figure \ref{fig:system_structure}, our system includes three sequential steps: (1) capturing facial gesture using FaceOSC, (2) connecting to compound facial expressions of emotion, and (3) synthesizing musical features based on the emotional state. FaceOSC software is used to help the Mugeetion system understand the user's facial gestures and generate sound based on the user's emotional state. The emotion detection module uses the software for real-time facial gesture tracking and transmits raw-level facial data over the Open Sound Control (OSC) protocol. If the detector finds multiple potential faces within the frame, the closest face will get the priority of recognition, analyzing a single face at a time. For analyzing facial expression, we use the FACS and Action Unit classification\footnote{Description of Facial Action Coding System and Action Units \url{https://www.cs.cmu.edu/~face/facs.htm}}. We chose the Action Unit (AU) combinations of three basic emotions: (A) happy, (B) neutral, and (C) sad. Each emotional state is combined with several individual AUs. For example, the facial expression of happy includes AU 6 (cheek raiser), AU 12 (lip corner puller), and AU 25 (lips part) (See Figure \ref{fig:system_structure}).

\begin{figure}[h]
\centering
\includegraphics[width=8cm, height=3.3cm]{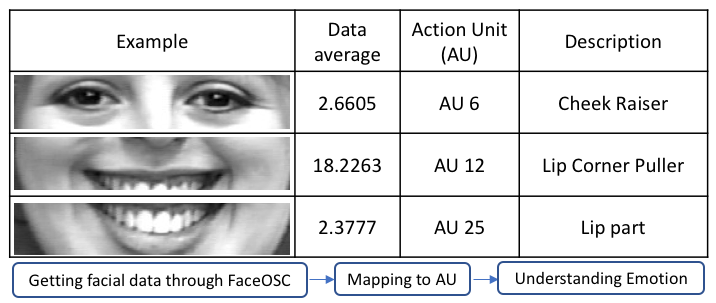}
  \caption{Data transition process: from facial gesture to emotion}
  \label{fig:AU_step}
\end{figure}

We practiced our sonification method with face images from The Cohn-Kanade AU-Coded Facial Expression Data-base \cite{kanade2000comprehensive,lucey2010extended}. By training with multiple images, we made the system work well with different faces. We selected representative facial images for linking with our sound generation process. We used 20 images for each emotional state: happy, neutral, sad (60 images total). We measured these data to create a data range for each emotional state and defined the differences between each emotion. We manually annotated the range of facial gestures for mapping each muscle activation to AU components (See Table \ref{tab:faceoscdata})\footnote{The unit in this table is followed by FaceOSC data measurement.}. Figure \ref{fig:AU_step} shows the data range for AU components with our training images. For example, the average AU6 scale of the happy face is 2.6605, AU12 is 18.2263, and AU25 is 2.3777. After learning the range of AUs, the system can classify facial gestures to pre-defined states with each individual photo or real-time face input through the connected web-cam. Along with categorizing, the system attempts to translate the musical style based on the input of emotional states (See Figure \ref{fig:AU_step}).

% In Figure 1, it explains about how FaceOSC can process facial data from real image points, so I am interested in exploring some representative facial expression based on point distribution model. 
\begin{table}
\begin{center}
\begin{tabular}{|l|c|r|}
    \hline
    position & details & data range (min/max)\\ \hline

    {\multirow{2}{*}{mouth}} & width & 6.0244/19.2747\\ \cline{2-3}
    & height & 0.8893/3.0010\\
    \hline
    {\multirow{2}{*}{eyebrow}} & left & 6.7666/8.0714\\ \cline{2-3}
    & right & 6.6787/7.9785\\
    \hline
    {\multirow{2}{*}{eye}} & left & 2.4329/3.4357\\ \cline{2-3} 
    & right & 2.3950/3.3144\\
    \hline
    jaw & - & 18.9888/22.9718\\
    \hline
    nostrils & - & 5.6477/8.8061\\
    \hline
    
%     position & - & 205.7676/287.6654\\ 
%     \hline
%     scale & - & 6.3838/8.9855\\ 
%     \hline
%     orientation & - & -\\
%     \hline
\end{tabular}    
\caption{Facial data configuration from FaceOSC}
\label{tab:faceoscdata}  
\end{center}
\end{table}

\subsection{Sonification with Action Units}
In this section, we focus on sonification with AUs in detail. We generate musical output based on the connection between AUs and emotional state. We then apply the formula between emotion and sound features, such as how the energetic happy face is mapped to the pitch/loudness increasing, and the dynamics in the sad face are mapped to white noise/distortion parameters. We also connect specific AUs to MIDI notes for sonification. The MIDI packets are mapped to controls of different parameters, resulting in different musical sounds based on how the emotional state moves. For example, when a user moves their mouth, the mouth height data is inputted and we normalize the data between 0-127 scales for generating MIDI notes or dynamics. Then, these 0-127 scales correspond to MIDI note scales. There are a few studies have explored this method before \cite{d2013mageface,migicovsky2014moveosc,sedes2004visualization}, and we explore the linkage between other sound features and the emotion conveyed in the AUs.

\subsection{Connecting between Emotion and Sound}
In this section, we explain how the system has been implemented for connecting emotion to sound. The system can interpolate the sound results from facial gesture inputs. In this approach, we generate the sound based on pre-recorded sound. We can play different sounds based on the user's happy, neutral, or sad emotional state. Our Mugeetion interface automatically plays the specific song related to the user's emotional state. We list our sound files below, which have been played to a number of subjects interacting with the system.

\begin{description}
\item [$\cdot$] Happy \\ Mozart - The Piano Sonata No 16 in C major \\ Mozart - Eine Kleine Nachtmusik K 525 Allegro
\vspace*{-3mm}
\item [$\cdot$] Neutral \\ Mozart - Piano Sonata No 11 in A major K 331
\vspace*{-3mm}
\item [$\cdot$] Sad \\ Mozart - Symphony No 25 in G Minor K 183 \\1st Movement \\ Mozart - Requiem in D minor 
% \vspace*{-3mm}
% \item [$\cdot$] (Optional) Surprise \\ Mozart - Die Entfuhrung aus dem Serail K 384 Turkish Finale
\end{description}

The selection of the list is based on the study of the Mozart Effect \cite{perlovsky2013mozart}. For the sound files, we use Piano-midi.de dataset\footnote{\url{http://www.piano-midi.de/mozart.htm}}.

% Moreover, every AU per second will automatically save into the log file with its sound output also. Using the log file, we want to see how the results would change if we want to train on the combinations of AUs and musical output. First, the system would be able to store a collection of data that causes the user to satisfy their musical output, which creators can use to bolster their sonification process. Further, we expect that the method would facilitate deliberately recalling the memory that occurred throughout the mixing process, which can reuse the moment for generating the patterns in their musical output.

\section{Prototypes}  
\subsection{Demo}

For the prototype of our system, we explored adding more musical variation, such as pitch height, loudness, distortion, or tempo change, as parameters to be controlled. We show a possibility of sound generation in real-time. %In addition, we would design filtering object for magnifying specific facial data, which can result in novel sonification. 
Our preliminary demo video is uploaded on Youtube\footnote{\url{https://www.youtube.com/playlist?list=PLjaQX_vKy2Jcv0r9wrc_yU2gbRhZh39GV}}. In order to allow users to easily interact with their sound generation process, we built a Max application, which utilizes facial data and FAC for sound creation.

\subsection{Sound Installation Work with Mugeetion}
Our sonification method, Mugeetion, has also been used in the sound installation exhibition, \textit{Hearing Seascape} (See Figure 3) at the Qualcomm Institute at UC San Diego in February 2018\footnote{Photo by Alex Matthews \raisebox{.5pt}{\textcircled{\raisebox{-.9pt} {c}}}2018 Regents of the University of California.}. This exhibition was a part of a collaborative effort with the Scripps Institution of Oceanography at UC San Diego to interpret their coral reef image data in a musical way. To convey the importance of engaging in the soundscapes of coral reefs, we suggested that our Mugeetion would be effective in fulfilling the goal of the project.
% In the installation work, based on the notion of acoustic ecology, we want to delve into the innovative design to bring out the positive aspects of sound in the ocean environment. %\footnote{http://ideas.calit2.net/performances.php}%Calit2 IDEAS exhibition 2018 
% This exhibition represented one of collaboration results from Calit2 Center of Graphics, Visualization and Virtual Reality, and Scripps Institution of Oceanography at UC San Diego.
Our prototype of the exhibition can be found in Youtube\footnote{\url{https://youtu.be/c-kHwnYuF44}}. The main goals of this project were to display different aspects of sound and innovative graphic design to create an enjoyable environment for the audience, and to create an inviting soundscape with a synergy among voices, images, synthesized sounds, and human emotion. 
\begin{figure}[h]
\centering
\includegraphics[width=4cm, height=3cm]{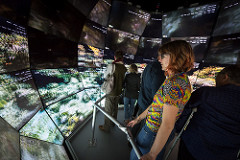}
\includegraphics[width=4cm, height=3cm]{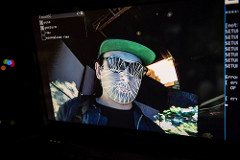}
  \caption{Left: \textit{Hearing Seascape} exhibition, Right: Interaction with Mugeetion during the exhibition (Neutral state)}
  \label{fig:hearingseascape}
\end{figure}

\subsubsection{Characteristics of the sounds in the Hearing Seascape}

There were two sound components in the sound installation. First, regarding sound input, we made recordings of singing and speaking in bowls of water. We recorded various sounds, such as giggling, clicking with tongue, singing, spoken dialogue, low/high pitches, both in the air and in the water. This specific sonification process was related to the goal of the project. The sound of voices underwater showed a variation of pitch and vagueness of speech. This is representative of the confusion and misunderstanding that surrounds coral reef research \cite{smith2016re}-- there is so much yet to be discovered and understood about these creatures. Second, using Mugeetion, our method detected the audience's facial expression in real-time, and the detected emotional state was used to display of the coral reef images and synthesize the soundscape. The audience can hear the sound that is simultaneously mapped with their specific facial gestures. For instance, when audiences expressed strong emotions with their facial gestures, these dynamics connected to sound components to increase intensity, tempo, and pitch height. The interaction through Mugeetion invited the audience to participate in the exhibition.
\\ 
%\footnote{This exhibition is collaborated with Lauren Jones who is currently pursuing her Master's degree in Vocal Performance at UC San Diego. 
% After recording, we have synthesized some sounds related to water in Max and mapped the sound with facial components. \\

\section{Discussion and Future Work}
Mugeetion makes several contributions to previous work. Rather than simply detecting facial gesture data, it also automatically extracts emotional states and produces sound output transition. Mugeetion provides a sound generation model to users based on the components of emotion and musical metadata. 
We focus on how sound can be changed based on users' emotional movement. In the presented soundscape installation, the interaction between emotion and sound occurred based on user's emotional states. We explore how audience participation in artwork can be utilized in interactive systems and how it changes the sound generation output. %Based on the motivation of each artwork, participants might be suggested to follow the guidance get the best possible experience. 
In future work, we will collect continuous auditory feedback during the exhibition in order to evaluate the sound generation output. For example, audiences would be asked how satisfied they were with the reflection between sound output and their emotional states. 

Furthermore, the system would be able to store a collection of data, which creators can use to improve their sonification process. Every AU per second and audio files would be automatically saved. The system would collect and store a repository of the memory units that users can look back on in order to re-utilize their composition process.
%It would record the content so that users can re-use the log data from the emotion input and sound output. 

We will further develop the system based on the following issues:
\begin{description}
\item [$\cdot$] increasing training images for covering multiple faces and optimizing different emotional states
\vspace*{-2mm}
% \item [$\cdot$] exploring the sonification objects for constructing concrete emotion-sound parameters
% \vspace*{-1mm}
\item [$\cdot$] implementing an AU indicator or other emotional measure on the FaceOSC display for better interaction with users
% \vspace*{-2mm}
% \item [$\cdot$] conducting user study in various Mugeetion settings 
\vspace*{-2mm}
\item [$\cdot$] exploring other similar emotion interactive system to compare the sonification result
\end{description}

\begin{acknowledgments}
This paper has been generously supported by Dr. Lei Liang, other members from Calit2 Center of Graphics, Visualization and Virtual Reality, and The Smith Lab in Scripps Institution of Oceanography at UC San Diego.
\end{acknowledgments}

\bibliography{icmc2018template}

\end{document}